\documentclass[runningheads]{llncs}
\usepackage{graphicx}

\usepackage{tikz}
\usepackage{comment}
\usepackage{amsmath,amssymb} 
\usepackage{color}

\usepackage[width=122mm,left=12mm,paperwidth=146mm,height=193mm,top=12mm,paperheight=217mm]{geometry}

\usepackage{multirow}
\usepackage{booktabs}

\usepackage{subfigure}

\newcommand{\etal}{\textit{et al}.}

\usepackage[capitalize]{cleveref}
\crefname{section}{Sec.}{Secs.}
\Crefname{section}{Section}{Sections}
\Crefname{table}{Table}{Tables}
\crefname{table}{Tab.}{Tabs.}

\begin{document}
\pagestyle{headings}
\mainmatter
\def\ECCVSubNumber{182}  

\title{Domain Adaptation via Bidirectional Cross-Attention Transformer} 

\author{Xiyu Wang, Pengxin Guo, Yu Zhang\thanks{Corresponding author}}
\institute{Department of Computer Science and Engineering\\Southern University of Science and Technology}

\maketitle

\begin{abstract}
Domain Adaptation (DA) aims to leverage the knowledge learned from a source domain with ample labeled data to a target domain with unlabeled data only. Most existing studies on DA contribute to learning domain-invariant feature representations for both domains by minimizing the domain gap based on convolution-based neural networks. Recently, vision transformers significantly improved performance in multiple vision tasks. Built on vision transformers, in this paper we propose a Bidirectional Cross-Attention Transformer (BCAT) for DA with the aim to improve the performance. In the proposed BCAT, the attention mechanism can extract implicit source and target mixup feature representations to narrow the domain discrepancy. Specifically, in BCAT, we design a weight-sharing quadruple-branch transformer with a bidirectional cross-attention mechanism to learn domain-invariant feature representations. Extensive experiments demonstrate that the proposed BCAT model achieves superior performance on four benchmark datasets over existing state-of-the-art DA methods that are based on convolutions or transformers.

\end{abstract}

\section{Introduction} \label{sec:introduction}

Deep Neural Networks (DNNs) have achieved remarkable success on a variety of computer vision problems~\cite{krizhevsky2012imagenet,he2016deep,dosovitskiy2020image,liu2021swin}. However, such achievement heavily relies on a large amount of labeled training data, which is difficult to obtain in many real-world applications. To handle the labeled data scarcity problem, Domain Adaptation (DA)~\cite{yang2020transfer} has been proposed to transfer the knowledge learned from a source domain with ample label data to help learn in a target domain with unlabeled data only. The core idea of DA is to learn a domain-invariant feature representation, which could be both transferable to narrow the domain discrepancy and discriminative for classification in the target domain. To achieve this goal, in past decades many DA methods have been proposed and they can be classified into two main categories: \textit{domain alignment} methods~\cite{tzeng2014deep,long2015learning,chen2019joint} and \textit{adversarial learning} methods~\cite{ganin2015unsupervised,wang2020transfer}.

With great success in Natural Language Processing (NLP)~\cite{vaswani2017attention}, recently transformer also has received increasing attention in the computer vision community, including Vision Transformer (ViT)~\cite{dosovitskiy2020image}, Data-efficient image Transformers (DeiT)~\cite{touvron2021training}, and Swin transformer (Swin)~\cite{liu2021swin}. Different from Convolutional Neural Networks (CNNs) that act on local receptive fields of images, transformers model long-range dependencies among visual features across the image through the self-attention mechanism. Due to its advantages in context modeling, vision transformers have obtained excellent performance on various vision tasks, such as image classification~\cite{liu2021swin,touvron2021training,han2021transformer}, object detection~\cite{carion2020end}, dense prediction~\cite{ranftl2021vision} and video understanding \cite{girdhar2019video,neimark2021video}.

There are some works \cite{yang2021transformer,xu2021cdtrans,yang2021tvt,munir2021sstn} to apply transformers to solve DA problems. Some works \cite{yang2021transformer,munir2021sstn,yang2021tvt} directly apply vision transformers but ignore the property of DA problems. To make the vision transformers more suitable for DA tasks, Xu \etal~\cite{xu2021cdtrans} propose a Cross-Domain Transformer (CDTrans) which consists of a weight-sharing triple-branch transformer to utilize the self-attention and cross-attention mechanisms for both feature learning and domain alignment. However, the CDTrans model only considers one-directional cross-attention from the source domain to the target domain but ignores the cross-attention from the target domain to the source domain. Furthermore, during the training process, the CDTrans model restricts the data in a mini-batch to be source and target images from the same class, which brings additional difficulties to accurately determine pseudo labels for unlabeled target data and restricts its applications.

To remedy those limitations, we propose a Bidirectional Cross-Attention Transformer (BCAT) to help appropriately learn domain-invariant feature representations. In BCAT, we construct the bidirectional cross-attention to enhance the transferability of vision transformers. The bidirectional cross-attention naturally fits knowledge transfer between the source and target domains in both directions, and it enables the implicit feature mixup between domains. The proposed BCAT model combines the bidirectional cross-attention with the self-attention as quadruple transformer blocks for learning two augmented feature representations. The quadruple transformer blocks can holistically focus on intra- and inter-domain features and blur the boundary between the two domains. By minimizing the Maximum Mean Discrepancy (MMD) \cite{gretton2006kernel} between the learned feature representations in both domains, the BCAT could decrease the domain gap and learn domain-invariant feature representations.

In summary, our contributions are three-fold.
\begin{itemize}

\item We propose a quadruple transformer block to combine both self-attention and cross-attention to learn augmented feature representations for both source and target domains.

\item Built on the quadruple transformer block, we propose the BCAT under the DA setting to learn domain-invariant feature representations.

\item The proposed BCAT outperforms state-of-the-art baseline methods on four benchmark datasets.

\end{itemize}

\section{Related Work} \label{sec:rw}


\paragraph{Vision Transformers.} Transformer is first proposed in~\cite{vaswani2017attention} to model sequential text data in the NLP field. Dosovitskiy \etal~\cite{dosovitskiy2020image} firstly apply transformers to computer vision and propose ViT by feeding transformers with sequences of image patches. Then, many variants of ViT~\cite{touvron2021training,han2021transformer,ranftl2021vision,liu2021swin} are proposed to achieve promising performance on computer vision tasks when compared with their CNN counterparts. Liu \etal~\cite{liu2021swin} propose the Swin transformer, which performs local attention within a window and introduces a shifted window partitioning approach for cross-window connections.

\paragraph{Vision Transformer for Domain Adaptation.} There are some works to apply vision transformers to solve DA problems. For example, Yang \etal~\cite{yang2021transformer} incorporate the transformer into a CNN to focus on essential regions. Xu \etal~\cite{xu2021cdtrans} propose a weight-sharing triple-branch transformer to utilize the self-attention and cross-attention mechanisms for both feature learning and domain alignment. Yang \etal~\cite{yang2021tvt} design a Transferable Vision Transformer (TVT), which can enforce ViT to focus on both transferable and discriminative features by injecting learned transferability into attention blocks. 

\section{The BCAT Method}

In this section, we introduce the proposed BCAT method.

For a DA problem, we are given a labeled source dataset $\mathcal{D}_s = \{ (x_s^i, y_s^i) \}_{i=1}^{n_s}$ and an unlabeled target dataset $\mathcal{D}_t = \{x_t^i\}_{i=1}^{n_t}$, where $n_s$ and $n_t$ denote the number of instances in the source and target domains, respectively. These two domains have different data distributions, i.e., $p_s(x_s) \neq p_t(x_t)$, due to the domain shift, but they share an identical label space, i.e., $\mathcal{Y}_t = \mathcal{Y}_s$. The goal of DA problem is to train a model that can utilize the useful knowledge in the source domain $\mathcal{D}_s$ to help the learning in the target domain $\mathcal{D}_t$.

\subsection{Quadruple Transformer Block 
}

\begin{figure}[t]
\centering
\includegraphics[width=\linewidth]{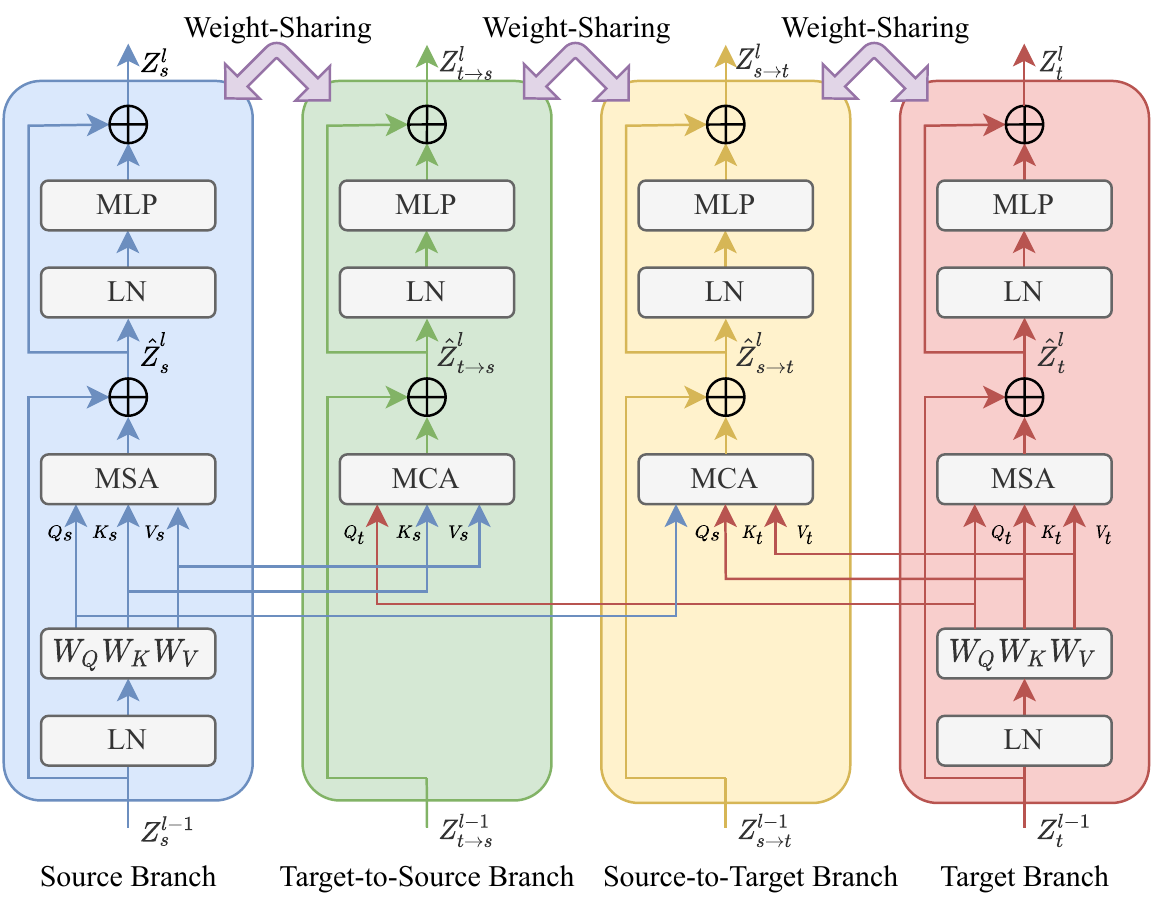}
\caption{A quadruple transformer block with two self-attention modules and a bidirectional cross-attention module. `MSA' and `MCA' denote the multi-head self-attention and multi-head cross-attention, respectively. `LN' denotes the layer normalization. `MLP' denotes a multi-layer perceptron. $W_Q$, $W_K$, and $W_V$ denote transformation parameters to generate queries, keys, and values, respectively.}
\label{fig:cross_attention}
\end{figure}

As shown in Figure~\ref{fig:cross_attention}, the BCAT method combines two cross-attention modules with two self-attention modules to design a quadruple transformer block.

Specifically, for the self-attention, $N$ image patches are first transformed into queries $Q \in \mathbb{R}^{N \times d_{k}}$, keys $K \in \mathbb{R}^{N \times d_{k}}$, and values $V \in \mathbb{R}^{N \times d_{v}}$, where $d_k$ and $d_v$ indicate their dimensions. The queries and the corresponding keys calculate weights assigned to each value by a compatibility function, while a weighted sum of the values is the output of self-attention. The self-attention for image patches $\mathbf{x}$ can be computed as
\begin{equation} \label{eq:atten_self}
\text{Attn}_{\text{self}} (\mathbf{x}) = \text{softmax} \left(\frac{QK^{T}}{\sqrt{d_{k}}}\right) V ,
\end{equation}
where $\text{softmax}(\cdot)$ denotes the softmax function.

Different from the self-attention module, the cross-attention module computes pairwise weights between different images. We leverage the cross-attention module to produce mixup features to blur the boundary between the source and target domain. Specifically, we define bidirectional cross-attention as bilaterally executing the cross-attention between source and target domains as shown in the middle of Figure \ref{fig:cross_attention}, and it is formulated as
\begin{alignat}{3}
\text{Attn}_{\text{bidir}} (x_s, x_t) &= \text{Attn}_{\text{cross}} (x_t, x_s) + \text{Attn}_{\text{cross}} (x_s, x_t), \nonumber\\
\text{Attn}_{\text{cross}} (x_t, x_s) &= \text{softmax} \left(\frac{Q_t K_s^{T}}{\sqrt{d_{k}}}\right) V_s, \label{eq:atten_cross_1} \\
\text{Attn}_{\text{cross}} (x_s, x_t) &= \text{softmax} \left(\frac{Q_s K_t^{T}}{\sqrt{d_{k}}}\right) V_t, \label{eq:atten_cross_2}
\end{alignat}
where $x_s$ denotes patches of a source image, $x_t$ denotes patches of a target image, and $Q$, $K$, and $V$ with subscripts $s$ and $t$ denote queries, keys, and values based on the source and target image patches, respectively.

With two self-attention modules and the bidirectional cross-attention module as shown in Figure~\ref{fig:cross_attention}, there are four weight-sharing transformers in the quadruple transformer block. We respectively name these four branches from left to right as source branch, target-to-source branch, source-to-target branch, and target branch. In the source and target branches, source image patches, and target image patches independently serve as the inputs for transformer blocks with Multi-head Self-Attention (MSA) modules and Multi-Layer Perceptron (MLP) modules to extract the source feature representation $Z_s$ and the target feature representation $Z_t$, respectively. Different from the source and target branches, the source-to-target branch, and the target-to-source branch apply the bidirectional cross-attention modules to learn intermediate feature representations between the source and target domain in two directions. A quadruple transformer block is mathematically formulated as
\begin{alignat*}{8}
\hat{Z}_s^l &= \text{MSA} \left( \text{LN} (Z_s^{l-1}) \right) + Z_s^{l-1}, \\
Z_s^l &= \text{MLP} \left( \text{LN} (\hat{Z}_s^l) \right) + \hat{Z}_s^l, \\
\hat{Z}_{t \rightarrow s}^l &= \text{MCA} \left( \text{LN} (Z_t^{l-1}), \text{LN} (Z_s^{l-1})\right) + Z_{t \rightarrow s}^{l-1}, \\
Z_{t \rightarrow s}^l &= \text{MLP} \left( \text{LN} (\hat{Z}_{t \rightarrow s}^l) \right) + \hat{Z}_{t \rightarrow s}^l,\\
\hat{Z}_{s \rightarrow t}^l &= \text{MCA} \left( \text{LN} (Z_s^{l-1}), \text{LN} (Z_t^{l-1}) \right) + Z_{s \rightarrow t}^{l-1}, \\
Z_{s \rightarrow t}^l &= \text{MLP} \left( \text{LN} (\hat{Z}_{s \rightarrow t}^l) \right) + \hat{Z}_{s \rightarrow t}^l,\\
\hat{Z}_t^l &= \text{MSA} \left( \text{LN} (Z_t^{l-1}) \right) + Z_t^{l-1}, \\
Z_t^l &= \text{MLP} \left( \text{LN} (\hat{Z}_t^l) \right) + \hat{Z}_t^l,
\end{alignat*}
where $Z_s^{l-1}$, $Z_{t \rightarrow s}^{l-1}$, $Z_{s \rightarrow t}^{l-1}$ and $Z_t^{l-1}$ denote inputs for the $l$th quadruple transformer block, $Z_s^l$, $Z_{t \rightarrow s}^l$, $Z_{s \rightarrow t}^l$ and $Z_t^l$ denote the corresponding outputs for the $l$th quadruple transformer block, $\text{LN}(\cdot)$ denotes the layer normalization \cite{wang2019learning}, $\text{MLP}$ is 2-layer fully connected neural network with the GELU activation function  \cite{hendrycks2016gaussian}, and $\text{MSA}(\cdot)$ and $\text{MCA}(\cdot)$ denote multi-head self-attention and multi-head cross-attention, respectively. Here $\text{MSA}(\cdot)$ relies on $\text{Attn}_{\text{self}}(\cdot)$ defined above but with multiple attention heads and $\text{MCA}(\cdot)$ is defined similarly based on $\text{Attn}_{\text{cross}}(\cdot,\cdot)$. $Z_s^0$ and $ Z_{t \rightarrow s}^{0}$ will be initialized to raw source image patches, while $Z_t^0$ and $ Z_{s \rightarrow t}^{0}$ will be initialized to raw target image patches.

We use two combined feature representations $[Z_s,Z_{t \rightarrow s}]$ and $[Z_t,Z_{s \rightarrow t}]$ as the augmented feature representations for the source and target domains, respectively. The representation $[Z_s,Z_{t \rightarrow s}]$ is regarded as a source-dominant feature representation, while the augmented feature representation $[Z_t,Z_{s \rightarrow t}]$ can be viewed as the target-dominant feature representation.

\subsection{Bidirectional Cross-Attention as Implicit Feature Mixup}

\begin{figure*}[htbp]
\centering
\includegraphics[width=1\textwidth]{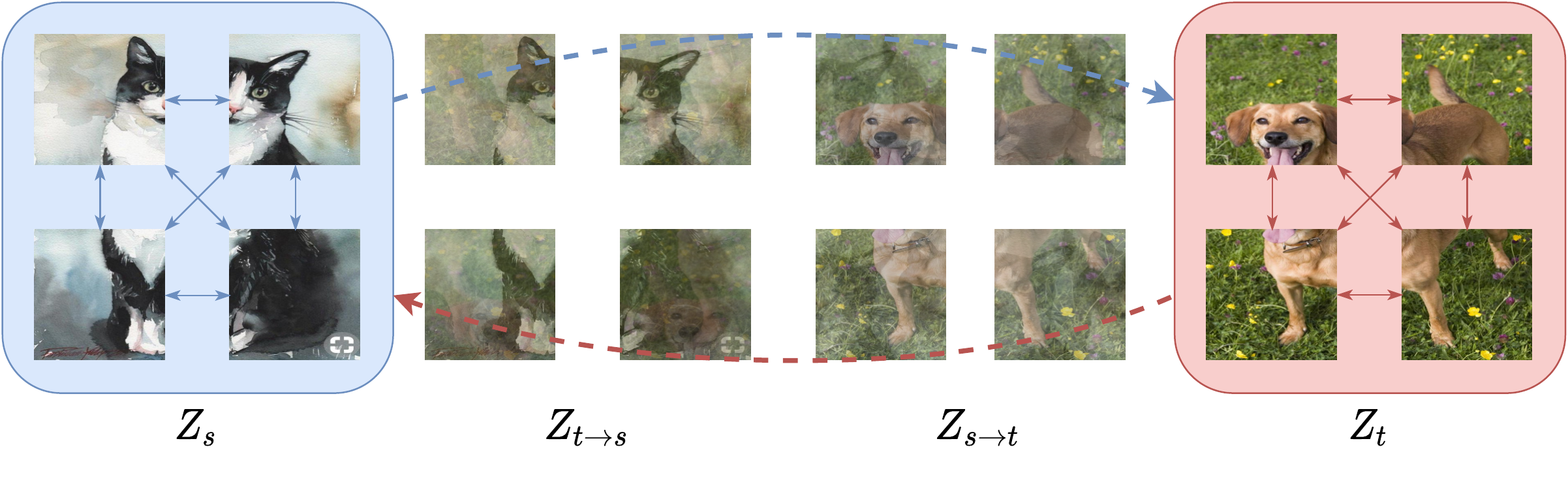}
\caption{An illustration of the bidirectional cross-attention mechanism to perform implicit feature mixup.}
\label{fig:mixup}
\end{figure*}

To see why the bidirectional cross-attention mechanism used in the proposed quadruple transformer block works for DA, we use Figure \ref{fig:mixup} as an illustration. In Figure \ref{fig:mixup}, $Z_s$ and $Z_t$ seem totally different, which poses challenges for DA. The bidirectional cross-attention mechanism generates $Z_{t \rightarrow s}$ and $Z_{s \rightarrow t}$. According to Figure \ref{fig:mixup}, $Z_{t \rightarrow s}$ and $Z_{s \rightarrow t}$ could be viewed as the mixup of $Z_s$ and $Z_t$ but with different proportions. Compared with the direct mixup on images from source and target domain, each image patch is weighted by all image patches from another domain,
which makes such mixup more smooth and robust. It is easy to see that $Z_{t \rightarrow s}$ is more close to $Z_s$ but $Z_{s \rightarrow t}$ is more close to $Z_t$. Since the discrepancy between $Z_{t \rightarrow s}$ and $Z_{s \rightarrow t}$ is smaller than that between $Z_s$ and $Z_t$, $Z_{t \rightarrow s}$ and $Z_{s \rightarrow t}$ could be used as a bridge to connect $Z_s$ and $Z_t$ to help minimize the domain discrepancy. Hence, minimizing the domain discrepancy between the source-dominant feature representation (i.e., $[Z_s,Z_{t \rightarrow s}]$) and the target-dominant feature representation (i.e., $[Z_t,Z_{s \rightarrow t}]$) could be easier than directly minimizing the source and target feature representations (i.e., $Z_s$ and $Z_t$). In this sense, the bidirectional cross-attention mechanism used in the proposed quadruple transformer block could decrease the domain discrepancy and help conduct the domain adaptation. In Section \ref{sec:ablation_study}, we could find empirical supports for such claim.


\subsection{Loss Functions}

\begin{figure*}[htbp]
\centering
\includegraphics[width=\textwidth]{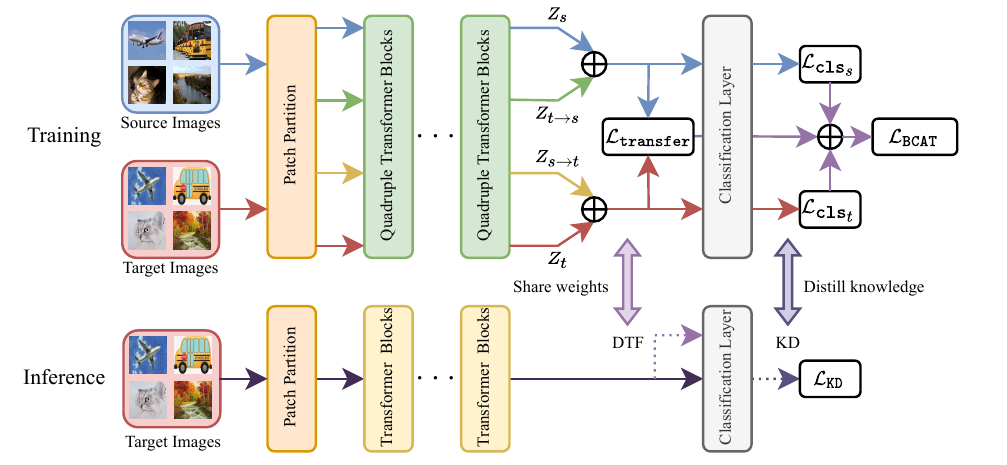}
\caption{Top: the architecture of the proposed BCAT method. Bottom: the two inference models introduced in Section \ref{sec:inference}. }
\label{fig:bca}
\end{figure*}

For a DA problem, in this section, we introduce how to utilize quadruple transformer blocks in the proposed BCAT method and how to construct the objective function.


As shown in the top of Figure \ref{fig:bca}, the overall architecture of the BCAT method consists of a patch partition layer, multiple quadruple transformer blocks, and the classification layer. The patch partition layer is responsible of dividing an image $I\in \mathbb{R}^{H \times W \times 3}$ into a sequence of flattened 2D patches $\mathbf{x} \in \mathbb{R}^{N \times 3P^{2}}$, where $(P, P)$ is the size of each split image patch, and $N = HW/P^2$ is the total number of patches. Multiple quadruple transformer blocks are stacked together to learn augmented feature representations for both domains. The classification layer is to do the classification given the output of the last quadruple transformer block. For notation simplicity, in the following formulations, we assume that there is only one quadruple transformer block.

To align the source-dominant and target-dominant features in two domains, MMD is utilized to define the transfer loss as
\begin{align*}
\mathcal{L}_{\texttt{transfer}} = & \mathrm{MMD}^{2} ([G_{s}(x_s),G_{c}(x_t,x_s)], [G_{s}(x_t),G_{c}(x_s,x_t)] )
\label{eq:transfer_loss}
\end{align*}
where $x_s$ and $x_s$ represent the source and target data, $G_{s}(\cdot)$ denotes the output of the source or target branch in the quadruple transformer block due to the weight sharing mechanism, and $G_{c}(\cdot,\cdot)$ denotes the output of the target-to-source or source-to-target branch in the quadruple transformer block since these two branches share weights.

The classification layer is implemented as a 2-layer fully connected neural network. For the labeled source domain, we adopt the cross-entropy loss, which is defined as
\begin{equation*}
\mathcal{L}_{\texttt{cls}_s} = \mathrm{CE}(F([G_{s}(x_s),G_{c}(x_t,x_s)]),y_s),
\end{equation*}
where $F(\cdot)$ denotes the classification layer and $\mathrm{CE}(\cdot,\cdot)$ denotes the cross-entropy loss.
For unlabeled target data, we assign each instance a pseudo label in a way similar to the \mbox{ATDOC} method \cite{liang2021domain}.
Specifically, we use the average of predicted class probabilities of nearest neighbors for a target instance as its predicted class probability and choose the class with the largest probability as its pseudo label.
Then the classification loss on the target domain is formulated as
\begin{equation*}
\mathcal{L}_{\texttt{cls}_t} = 
-\frac{1}{N_t} \sum\nolimits_{i=1}^{N_t} \hat{q}_{i,\hat{y}_i} \log p_{i,\hat{y}_i},
\end{equation*}
where $\hat{y}_i=\arg\max_j\hat{q}_{i,j}$ denotes the assigned pseudo label for the corresponding target instance, and $p_i=F([G_{s}(x_t),G_{c}(x_s,x_t)])$ denotes the $K$-dimensional prediction with $K$ as the number of classes and $p_{i,\hat{y}_i}$ as its entry in the $\hat{y}_i$th dimension.
By combining those two loss functions, the total classification loss is formulated as
\begin{equation*}
\mathcal{L}_{\texttt{cls}} = \mathcal{L}_{\texttt{cls}_s} + \epsilon \mathcal{L}_{\texttt{cls}_t},
\end{equation*}
where $\epsilon$ is the ratio of the current epoch over the total number of epochs to reduce the effect of unreliable pseudo labels produced during the early training stage.

The total loss function of the BCAT method is formulated as
\begin{alignat}{1}
\mathcal{L}_\texttt{BCAT} = \mathcal{L}_{\texttt{cls}} + \beta \mathcal{L}_{\texttt{transfer}},
\end{alignat}
where
$\beta$ is a hyper-parameter.

\subsection{Inference for Target Domain}
\label{sec:inference}

During the inference process for the target domain, we can use the training model to make the prediction. However, this inference approach needs to utilize source data, which brings additional storage costs to access source data. In the following, we present two inference models, which do not require to utilize source data.

\paragraph{Knowledge Distillation (KD).} The first inference model is based on knowledge distillation \cite{hinton2015distilling}. Here the training model is used as a teacher, and an inference model with self-attention only is used as the student. We define the distillation loss with ``softmax-T'' as
\begin{equation*}
\mathcal{L}_{\texttt{KD}} = \alpha T^2  \sum\nolimits_N p_i \log q_i +\epsilon (1 - \alpha) \mathrm{CE}\left( F_S(G_{Ss}(x_t)), \hat{y}_{t} \right),
\end{equation*}
where $G_{Ss}(\cdot,\cdot)$ denotes the feature extractor with self-attention in the student model, $F_S(\cdot)$ denotes the classification layer in the student network, and $q_i$ and $p_i$ are ``softmax-T'' probabilities of logit outputs of the student network and teacher network, respectively, i.e., $\frac{\mathop{\exp}(z_i/T)}{\sum\nolimits_j \mathop{\exp}(z_j/T)}$.

\paragraph{Double Target Feature (DTF).} Another inference model for the target domain without using the source data is to use only the target branch in BCAT. Since we cannot compute the target-dominant feature representation without source data, we combine the target feature with itself as the target-dominant feature representation, i.e., $[Z_t,Z_t]$, as the input of the classification layer during the inference. Thus, the predicted output is formulated as $F\left(\left[G_{s}(x_t),G_{s}(x_t)\right]\right)$.

The above two inference models have no requirement to access source data during the inference. In experiments, we will compare those two inference models.

\section{Experiments}

In this section, we evaluate the proposed BCAT method.

\subsection{Setups}

\paragraph{Datasets.} We conduct experiments on four benchmark datasets, including Office-31 \cite{saenko2010adapting}, Office-Home \cite{venkateswara2017deep}, DomainNet \cite{peng2019moment}, and VisDA-2017 \cite{peng2017visda}. By following \cite{xu2021cdtrans}, we construct transfer tasks on those four datasets.
\begin{itemize}
\item The Office-31 dataset is a standard dataset for real-world domain adaptation, which includes three domains (i.e., Amazon (A), DSLR (D), and Webcam (W)) with 4,110 images of 31 categories.

\item The Office-Home dataset is a medium-sized benchmark than Office-31, which covers 65 classes with totally 15,550 images. It contains four different domains: Artistic images (Ar), Clip art (Cl), Product images (Pr), and Real-world images (Rw).

\item The DomainNet dataset is a more challenging large-sized benchmark with six distinct domains: real world images (R), sketch (S), painting artistic images (P), infographic images (I), clipart (C), and quickdraw (Q). Each domain includes 345 categories. We evaluated our method in six domain adaptation tasks among real, sketch and painting artistic domains.

\item The VisDA-2017 dataset is a large-sized database for simulation-to-real domain adaptation. The source domain includes approximately 155K synthetic images and the target domain consists of about 55K authentic images.
\end{itemize}

\paragraph{Baseline Methods.} We compare the proposed BCAT method with state-of-the-art DA methods on respective datasets, including Calibrated Multiple Uncertainties (CMU) \cite{fu2020learning}, Source HypOthesis Transfer (SHOT) \cite{liang2020we}, Domain Consensus Clustering (DCC) \cite{li2021domain}, Transferable Vision Transformer (TVT) \cite{yang2021tvt}, and Cross-Domain Transformer (CDTrans) \cite{xu2021cdtrans}, where TVT and CDTrans are based on transformers and other models are based on ResNet \cite{he2016deep}. We use 'Source-only' to denote a baseline model trained on source data only with its backbone depending on the context. The BCAT method with the KD inference model is denoted by BCAT-KD, and that with the DTF inference model is denoted by BCAT-DTF.

\paragraph{Implementation Details.}

For all the DA tasks, we use the ViT-B and Swin-B pretrained on the ImageNet dataset~\cite{deng2009imagenet} as the backbone network for the proposed BCAT method, and hence 12 quadruple transformer blocks for ViT-B and 24 quadruple transformer blocks for Swin-B are used in the BCAT method, respectively. For the BCAT method built on the ViT-B, we use the SGD method~\cite{robbins1951stochastic} with a momentum of 0.9 and a weight decay of $5\times 10^{-4}$ as the optimizer. We use a base learning rate of $8\times 10^{-3}$ for the Office-31, Office-Home, and DomainNet datasets and $8\times 10^{-4}$ for the VisDA-2017 dataset. For the BCAT method based on the Swin-B, the AdamW method~\cite{loshchilov2017decoupled} with a momentum of 0.9 and a weight decay of 0.05 is used as the optimizer. We use a base learning rate of $5\times 10^{-6}$ for the Office-31, Office-Home, and DomainNet datasets and $5\times 10^{-7}$ for the VisDA-2017 dataset. For all the datasets, we set the batch size to 64 and train the model in 20 epochs. $\alpha$, $\beta$, and $T$ in the proposed BCAT method are set to $0.8$, $3$, and $2$, respectively, for all the DA tasks.

\subsection{Results}

According to results shown in Tables \ref{tab:office31}-\ref{tab:VisDA}, we can see that transformer-based DA models perform better than ResNet-based DA models, which demonstrates that transformers have more powerful capacities than ResNet. In some datasets (e.g., Office-31, Office-Home, and DomainNet), we can see that transformer-based `Source-only' model performs comparable or even better than state-of-the-art ResNet-based DA models. Those results show the superiority of vision transformers over ResNet for DA tasks. 

\begin{table}[!htpb]
\caption{Accuracy (\%) on the Office-31 dataset. The best results built on the ResNet and ViT are marked in box and the best results built on all the architectures (i.e., ResNet, ViT, and Swin) are shown in bold.}
\centering
\resizebox{0.7\linewidth}{!}{
\begin{tabular}{l|c|cccccc|c}
\toprule
 Method & & A$\rightarrow$W & D$\rightarrow$W & W$\rightarrow$D & A$\rightarrow$D & D$\rightarrow$A & W$\rightarrow$A & Avg \\
\midrule
Source-only & \multirow{3}{*}{\rotatebox{90}{ResNet}} & 68.4 & 96.7 & 99.3 & 68.9 & 62.5 & 60.7 & 76.1 \\
SHOT & & 90.1 & 98.4 & 99.9 & 94.0 & 74.7 & 74.3 & 88.6 \\
FixBi & & 96.1 & 99.3 & 100.0 & 95.0 &78.7 & 79.4 & 91.4 \\
\midrule
Source-only & \multirow{5}{*}{\rotatebox{90}{ViT}} & 89.2 & 98.9 & 100.0 & 88.8 & 80.1 & 79.8 & 89.5 \\
CDTrans & & 97.6 & 99.0 & 100.0 & 97.0 & 81.1 & 81.9 & 92.8 \\
TVT & & 96.4 & \framebox[1.1\width]{99.4} & 100.0 & 96.4 & 84.9 & \framebox[1.1\width]{86.1} & 93.8 \\
\textbf{BCAT-KD (ours)} & & \framebox[1.1\width]{96.9} & 98.7 & \framebox[1.1\width]{100.0} & \framebox[1.1\width]{97.5} & \framebox[1.1\width]{85.5} & 86.0 & \framebox[1.1\width]{94.1}  \\
\textbf{BCAT-DTF (ours)} & & 96.1 & 99.1 & \framebox[1.1\width]{100.0} & \framebox[1.1\width]{97.5} & 84.9 & 85.8 & 93.9  \\
\midrule
Source-only & \multirow{3}{*}{\rotatebox{90}{Swin}} & 89.2 & 94.1 & 100.0 & 93.1 & 80.9 & 81.3 & 89.8 \\
\textbf{BCAT-KD (ours)} & & \textbf{99.4} & \textbf{99.5} & \textbf{100.0} & \textbf{99.8} & \textbf{85.7} & \textbf{86.1} & \textbf{95.1} \\
\textbf{BCAT-DTF (ours)} & & 99.2 & \textbf{99.5} & \textbf{100.0} & 99.6 & \textbf{85.7} & \textbf{86.1} & 95.0 \\
\bottomrule
\end{tabular}
}
\label{tab:office31}
\end{table}

\paragraph{Office-31.} According to Table \ref{tab:office31}, the proposed ViT-based BCAT method outperforms the other DA method based on ViT, and it achieves the best average accuracy of 94.1\%. Moreover, on some transfer tasks (e.g., A $\rightarrow$ W, A $\rightarrow$ D, and D $\rightarrow$ A), the proposed BCAT method performs better than CDTrans and TVT, which demonstrates the effectiveness of the proposed BCAT method.

\begin{table*}[!htbp]
\caption{Accuracy (\%) on the Office-Home dataset. The best results built on the ResNet and ViT are marked in box and the best results built on all the architectures (i.e., ResNet, ViT, and Swin) are shown in bold.}
\centering
\resizebox{\linewidth}{!}{
\begin{tabular}{l|c|cccccccccccc|c}
\toprule
 Method & & Ar$\rightarrow$Cl & Ar$\rightarrow$Pr & Ar$\rightarrow$Rw & Cl$\rightarrow$Ar & Cl$\rightarrow$Pr & Cl$\rightarrow$Rw & Pr$\rightarrow$Ar & Pr$\rightarrow$Cl & Pr$\rightarrow$Rw & Rw$\rightarrow$Ar & Rw$\rightarrow$Cl & Rw$\rightarrow$Pr & Avg  \\
\midrule
Source-only & \multirow{3}{*}{\rotatebox{90}{ResNet}} & 34.9 & 50.0 & 58.0 & 37.4 & 41.9 & 46.2 & 38.5 & 31.2 & 60.4 & 53.9 & 41.2 & 59.9 & 46.1 \\
SHOT & & 57.1 & 78.1 & 81.5 & 68. & 78.2 &78.1 & 67.4 & 54.9 & 82.2 & 73.3 & 58.8 & 84.3 & 71.8 \\
FixBi & & 58.1 & 77.3 & 80.4 & 67.7 & 79.5 & 78.1 & 65.8 & 57.9 & 81.7 & 76.4 & 62.9 & 86.7 & 72.7 \\
\midrule
Source-only & \multirow{5}{*}{\rotatebox{90}{ViT}} & 66.2 & 84.3 & 86.6 & 77.9 & 83.3 & 84.3 & 76.0 & 62.7 & 88.7 & 80.1 & 66.2 & 88.6 & 78.8 \\
CDTrans & & 68.8 & 85.0 & 86.9 & 81.5 & 87.1 & 87.3 & 79.6 & 63.3 & 88.2 & 82.0 & 66.0 & 90.6 & 80.5 \\
TVT & & \framebox[1.1\width]{74.9} & 86.8 & 89.5 & 82.8 & 88.0 & 88.3 & 79.8 & 71.9 & 90.1 & 85.5 & 74.6 & 90.6 & 83.6 \\
\textbf{BCAT-KD (ours)} & & 74.6 & \framebox[1.1\width]{\textbf{90.8}} & \framebox[1.1\width]{90.9} & \framebox[1.1\width]{85.2} & \framebox[1.1\width]{\textbf{91.5}} & \framebox[1.1\width]{90.4} & \framebox[1.1\width]{84.5} & 74.3 & \framebox[1.1\width]{91.0} & 85.5 & \framebox[1.1\width]{74.8} & \framebox[1.1\width]{92.4} & \framebox[1.1\width]{85.5} \\
\textbf{BCAT-DTF (ours)} & & 74.2 & 90.6 & \framebox[1.1\width]{90.9} & 84.2 & 90.9 & 89.9 & 84.1 & \framebox[1.1\width]{\textbf{74.5}} & 90.8 & \framebox[1.1\width]{85.7} & \framebox[1.1\width]{74.8} & 92.2 & 85.2  \\
\midrule
Source-only & \multirow{3}{*}{\rotatebox{90}{Swin}} & 64.5 & 84.8 & 87.6 & 82.2 & 84.6 & 86.7 & 78.8 & 60.3 & 88.9 & 82.8 & 65.3 & 89.6 & 79.7  \\
\textbf{BCAT-KD (ours)} & & \textbf{75.4} & 90.0 & 92.8 & 88.0 & 90.4 & \textbf{92.8} & 87.1 & 74.1 & 92.4 & 86.2 & \textbf{75.8} & \textbf{93.5} & 86.5 \\
\textbf{BCAT-DTF (ours)} & & 75.3 & 90.0 & \textbf{92.9} & \textbf{88.6} & 90.3 & 92.7 & \textbf{87.4} & 73.7 & \textbf{92.5} & \textbf{86.7} & 75.4 & \textbf{93.5} & \textbf{86.6}  \\
\bottomrule
\end{tabular}
}
 \label{tab:office_home}
\end{table*}

\paragraph{Office-Home.} As shown in Table \ref{tab:office_home}, the proposed ViT-based BCAT method has the highest average accuracy of 85.5\%. Compared with the best performant baseline method (i.e., the TVT method), the proposed BCAT method significantly improves the performance in almost all the transfer tasks. Although transferring to the Cl domain is an arduous task as each method has the lowest average accuracy on such transfer tasks than other transfer tasks, we use the same settings of hyper-parameters as other transfer tasks to achieve comparable or better performance, which in some extent demonstrates the good generalization ability of the proposed BCAT method.

\paragraph{DomainNet.} According to the results on the DomainNet dataset shown in Table \ref{tab:domainnet}, the proposed ViT-based BCAT method achieves performance at a brand-new level with the average accuracy of 65.0\%. Especially for transfer tasks P$\rightarrow$R, R$\rightarrow$P,
R$\rightarrow$S, and
S$\rightarrow$R, the proposed BCAT method has a performance improvement of 10.6\%, 15.6\%, 19.2\%, and 11.6\%, respectively, over the CDTrans method that is also built on ViT. Results on this dataset again verify the effectiveness of the proposed BCAT method.

\begin{table}[!htpb]
\caption{Accuracy (\%) on the DomainNet dataset. The best results built on the ResNet and ViT are marked in box and the best results built on all the architectures (i.e., ResNet, ViT, and Swin) are shown in bold.}
\centering
\resizebox{0.7\linewidth}{!}{
\begin{tabular}{l|c|cccccc|c}
\toprule
 Method & & P$\rightarrow$R & R$\rightarrow$P & P$\rightarrow$S & S$\rightarrow$P & R$\rightarrow$S & S$\rightarrow$R & Avg \\
\midrule
Source-only & \multirow{3}{*}{\rotatebox{90}{ResNet}} & 30.1 & 28.3 & 27.0 & 27.0 & 26.9 & 29.7 & 28.2 \\
CMU & & 50.8 & 52.2 & 45.1 & 44.8 & 45.6 & 51.0 & 48.3 \\
DCC & & 56.9 & 50.3 & 43.7 & 44.9 & 43.3 & 56.2 & 49.2 \\
\midrule
Source Only & \multirow{4}{*}{\rotatebox{90}{ViT}} & 64.4 & 47.2 & 41.8 & 44.8 & 31.5 & 57.5 & 47.9 \\
CDTrans & & 69.8 & 47.8 & 49.6 & 54.6 & 33.5 & 68.0 & 53.9 \\
\textbf{BCAT-KD (ours)} & & \framebox[1.1\width]{80.4} & \framebox[1.1\width]{63.4} & 52.5 & 61.9 & 52.4 & \framebox[1.1\width]{79.6} & \framebox[1.1\width]{65.0}  \\
\textbf{BCAT-DTF (ours)} & & 79.2 & 62.8 & \framebox[1.1\width]{52.7} & \framebox[1.1\width]{62.2} & \framebox[1.1\width]{52.7} & 78.5 & 64.7 \\
\midrule
Source-only & \multirow{3}{*}{\rotatebox{90}{Swin}} & 72.7 & 60.2 & 47.4 & 53.2 & 50.0 & 66.7 & 58.4  \\
\textbf{BCAT-KD (ours)} & & \textbf{81.2} & 67.5 & 59.7 & 65.8 & \textbf{60.9} & 80.8 & 69.3 \\
\textbf{BCAT-DTF (ours)} & & 81.1 & \textbf{67.7} & \textbf{60.4} & \textbf{66.8} & 60.7 &\textbf{80.9} & \textbf{69.6}  \\
\bottomrule
\end{tabular}
}
\label{tab:domainnet}
\end{table}

\begin{table*}[!htbp]
\caption{Accuracy (\%) on the VisDA-2017 dataset. The best results built on the ResNet and ViT are marked in box and the best results built on all the architectures (i.e., ResNet, ViT, and Swin) are shown in bold.}
\centering
\resizebox{\linewidth}{!}{
\begin{tabular}{l|c|cccccccccccc|c}
\toprule
 Method & & aero & bicycle & bus & car & horse & knife & motor & person & plant & skate &  train & truck & Avg\\
\midrule
Source-only & \multirow{3}{*}{\rotatebox{90}{ResNet}} & 55.1 & 53.3 & 61.9 & 59.1 & 80.6 & 17.9 & 79.7 & 31.2 & 81.0 & 26.5 & 73.5 & 8.5 & 52.4 \\
SHOT & & 94.3 & 88.5 & 80.1 & 57.3 & 93.1 & 94.9 & 80.7 & 80.3 & 91.5 & 89.1 & 86.3 & 58.3 & 82.9 \\
FixBi & & 96.1 & 87.8 & \framebox[1.1\width]{\textbf{90.5}} & \framebox[1.1\width]{\textbf{90.3}} & 96.8 & 95.3 & 92.8 & \framebox[1.1\width]{\textbf{88.7}} & 97.2 & 94.2 & 90.9 & 25.7 & 87.2 \\
\midrule
Source-only & \multirow{5}{*}{\rotatebox{90}{ViT}} & 98.2 & 73.0 & 82.5 & 62.0 & 97.3 & 63.5 & \framebox[1.1\width]{96.5} & 29.8 & 68.7 & 86.7 & \framebox[1.1\width]{\textbf{96.7}} & 23.6 & 73.2 \\
TVT & & 92.9 & 85.6 & 77.5 & 60.5 & 93.6 & \framebox[1.1\width]{\textbf{98.2}} & 89.4 & 76.4 & 93.6 & 92.0 & 91.7 & 55.7 & 83.9\\
CDTrans & & 97.1 & 90.5 & 82.4 & 77.5 & 96.6 & 96.1 & 93.6 & 88.6 & \framebox[1.1\width]{\textbf{97.9}} & 86.9 & 90.3 & 62.8 & 88.4 \\
\textbf{BCAT-KD (ours)} & & \framebox[1.1\width]{99.0} & \framebox[1.1\width]{\textbf{92.5}} & 87.4 & 74.8 & \framebox[1.1\width]{98.2} & 98.5 & 93.6 & 67.7 & 89.0 & 96.4 & 95.8 & \framebox[1.1\width]{\textbf{69.8}} & \framebox[1.1\width]{88.5} \\
\textbf{BCAT-DTF (ours)} & & 98.9 & 91.3 & 87.4 & 73.8 & 97.9 & 98.1 & 94.2 & 64.0 & 88.9 & \framebox[1.1\width]{\textbf{97.4}} & 96.0 & 60.7 & 87.4 \\
\midrule
Source-only & \multirow{3}{*}{\rotatebox{90}{Swin}} & 98.7 & 63.0 & 86.7 & 68.5 & 94.6 & 59.4 & \textbf{98.0} & 22.0 & 81.9 & 91.4 & \textbf{96.7} & 25.7 & 73.9 \\
\textbf{BCAT-KD (ours)} & & \textbf{99.1} & 91.5 & 86.8 & 72.4 & 98.6 & 98.1 & 96.5 & 82.1 & 94.4 & 96.0 & 93.9 & 61.1 & \textbf{89.2} \\
\textbf{BCAT-DTF (ours)} & & \textbf{99.1} & 91.6 & 86.6 & 72.3 & \textbf{98.7} & 97.9 & 96.5 & 82.3 & 94.2 & 96.0 & 93.9 & 61.3 & \textbf{89.2}\\
\bottomrule
\end{tabular}
}
\label{tab:VisDA}
\end{table*}

\paragraph{VisDA-2017.} As shown in Table \ref{tab:VisDA}, the performance of the proposed ViT-based BCAT method on the VisDA-2017 dataset achieves the best average accuracy of 88.5\%. Moreover, the proposed BCAT method achieves the best performance on five classes, including `aeroplane', `bicycle', `horse', `skate', and `truck', than other ViT-based methods such as CDTrans and TVT.

\subsection{BCAT for Other Vision Transformers}

The proposed BCAT method is applicable to other vision transformers than ViT, and in Tables \ref{tab:office31}-\ref{tab:VisDA} we show the performance of the BCAT method built on the Swin transformer. According to the results, we can see that the proposed BCAT method built on the Swin transformer has better performance than ViT-based DA methods and achieves the best average accuracy of 95.1\%, 86.6\%, 69.6\%, and 89.2\% on Office-31, Office-Home, DomainNet and VisDA-2017 dataset, respectively. Those results imply that the proposed BCAT method not only takes effect on the global attention mechanism as in ViT but also works on the local attention mechanism as in the Swin transformer.
Moreover, the Swin-based BCAT method has a larger improvement over its `Source-only' counterpart than CDTrans and TVT, which demonstrates the effectiveness of the BCAT method.

\subsection{Ablation Study}
\label{sec:ablation_study}

In Table \ref{tab:ablation}, we conduct ablation studies to study the effects of three loss functions: $\mathcal{L}_{\texttt{cls}_s}$, $\mathcal{L}_{\texttt{cls}_t}$, and $\mathcal{L}_{\texttt{transfer}}$, on the Office-Home dataset. For each loss function, we consider two choices: using only the self-attention (denoted by `self') , and using both the self-attention and cross-attention (denoted by `cross') during the feature extraction process. In this study, the BCAT-KD inference model is used.

\begin{table*}[!htbp]
\caption{Ablation study on the Office-Home dataset in terms of the accuracy (\%) based on the ViT and Swin transformer.}
\centering
\resizebox{\linewidth}{!}{
\begin{tabular}{lll|l|cccccccccccc|c}
\toprule
$\mathcal{L}_{\texttt{cls}_s}$ & $\mathcal{L}_{\texttt{cls}_t}$ & $\mathcal{L}_{\texttt{transfer}}$ & & Ar$\rightarrow$Cl & Ar$\rightarrow$Pr & Ar$\rightarrow$Rw & Cl$\rightarrow$Ar & Cl$\rightarrow$Pr & Cl$\rightarrow$Rw & Pr$\rightarrow$Ar & Pr$\rightarrow$Cl & Pr$\rightarrow$Rw & Rw$\rightarrow$Ar & Rw$\rightarrow$Cl & Rw$\rightarrow$Pr & Avg \\
\midrule
slef & - & - &  \multirow{6}{*}{\rotatebox{90}{ViT}} & 66.2 & 84.3 & 86.6 & 77.9 & 83.3 & 84.3 & 76.0 & 62.7 & 88.7 & 80.1 & 66.2 & 88.6 & 78.8 \\
slef & slef & slef & & 73.5 & 90.4 & 89.9 & 84.6 & 89.7 & 90.7 & 83.4 & 73.0 & 90.9 & 85.8 & 73.9 & 91.7 & 84.8  \\
cross & - & - &  & 66.9 & 87.1 & 89.8 & 82.7 & 89.1 & 89.5 & 79.4 & 65.4 & 90.4 & 82.4 & 65.0 & 91.0 & 81.5  \\
cross & cross & - & & 73.8 & 91.2 & 90.4 & 85.4 & 90.4 & 90.2 & 84.3 & 73.5 & 90.3 & 84.5 & 74.2 & 92.1 & 85.0  \\
cross & - & cross & & 72.1 & 87.2 & 89.4 & 83.7 & 89.3 & 89.4 & 81.8 & 72.7 & 90.0 & 84.0 & 73.0 & 91.1 & 83.6  \\
cross & cross & cross & & 74.6 & 90.8 & 90.9 & 85.2 & 91.5 & 90.4 & 84.5 & 74.3 & 91.0 & 85.5 & 74.8 & 92.4 & 85.5 \\
\midrule
slef & - & - &  \multirow{6}{*}{\rotatebox{90}{Swin}} & 64.5 & 84.8 & 87.6 & 82.2 & 84.6 & 86.7 & 78.8 & 60.3 & 88.9 & 82.8 & 65.3 & 89.6 & 79.7   \\
slef & slef & slef & & 72.6 & 90.0 & 91.8 & 86.9 & 89.6 & 91.8 & 84.9 & 71.9 & 92.2 & 87.6 & 74.3 & 92.9 & 85.5 \\
cross & - & - & & 67.9 & 84.0 & 88.0 & 82.9 & 85.1 & 86.5 & 76.0 & 64.3 & 87.9 & 81.9 & 66.1 & 89.8 & 80.0  \\
cross & cross & - & & 71.5 & 90.1 & 92.6 & 87.4 & 91.2 & 93.2 & 85.4 & 72.1 & 92.4 & 85.6 & 74.1 & 92.7 & 85.7 \\
cross & - & cross & & 68.3 & 85.3 & 89.6 & 84.5 & 85.4 & 88.2 & 83.5 & 66.8 & 89.8 & 85.9 & 70.4 & 90.8 & 82.4 \\
cross & cross & cross & & 75.4 & 90.1 & 92.8 & 88.0 & 90.4 & 92.8 & 87.1 & 74.1 & 92.4 & 86.2 & 75.8 & 93.5 & 86.5 \\
\bottomrule
\end{tabular}
}
 \label{tab:ablation}
\end{table*}

Comparing the first and second rows in Table \ref{tab:ablation}, we can see that both the pseudo labeling and MMD are effective for transformer-based backbones in DA tasks. The comparison between the fourth and fifth rows shows that the MMD and pseudo labeling method still work for cross-attention based feature augmentation. According to the first and third rows, the bidirectional cross-attention shows better performance than self-attention, which could support the usefulness of the proposed bidirectional cross-attention which is to conduct the implicit feature mixup. Compared with the second row, the proposed BCAT method corresponding to the last row performs better, which demonstrates the effectiveness of the bidirectional cross-attention mechanism used in BCAT.

\subsection{Comparison between Inference Models}

\begin{table}[!htbp]
\caption{Accuracy (\%), occupied GPU memory, and FPS on the Office-Home dataset for the three inference models.}
\centering
\resizebox{\linewidth}{!}{
\begin{tabular}{l|c|cccccccccccc|c|c|r}
\toprule
Method & & Ar$\rightarrow$Cl & Ar$\rightarrow$Pr & Ar$\rightarrow$Rw & Cl$\rightarrow$Ar & Cl$\rightarrow$Pr & Cl$\rightarrow$Rw & Pr$\rightarrow$Ar & Pr$\rightarrow$Cl & Pr$\rightarrow$Rw & Rw$\rightarrow$Ar & Rw$\rightarrow$Cl & Rw$\rightarrow$Pr & Avg & GPU Mem & FPS  \\
\midrule
\textbf{BCAT (ours)} & \multirow{3}{*}{\rotatebox{90}{ViT}} & 75.1 & 90.5 & 90.6 & 85.1 & 91.2 & 90.1 & 84.5 & 74.5 & 90.8 & 86.0 & 75.3 & 92.3 & 85.5 & 2241MB & 78.6 \\
\textbf{BCAT-KD (ours)} & & 74.6 & 90.8 & 90.9 & 85.2 & 91.5 & 90.4 & 84.5 & 74.3 & 91.0 & 85.5 & 74.8 & 92.4 & 85.5 & 1780MB & 276.2 \\
\textbf{BCAT-DTF (ours)} & & 74.2 & 90.6 & 90.9 & 84.2 & 90.9 & 89.9 & 84.1 & 74.5 & 90.8 & 85.7 & 74.8 & 92.2 & 85.2 & 1780MB & 267.2  \\
\midrule
\textbf{BCAT (ours)} & \multirow{3}{*}{\rotatebox{90}{Swin}} & 74.1 & 89.8 & 90.1 & 88.1 & 89.5 & 92.5 & 87.3 & 73.3 & 92.3 & 88.4 & 74.2 & 92.5 & 86.0 & 4087MB & 74.2 \\
\textbf{BCAT-KD (ours)} & & 75.4 & 90.0 & 92.8 & 88.0 & 90.4 & 92.8 & 87.1 & 74.1 & 92.4 & 86.2 & 75.8 & 93.5 & 86.5 & 2871MB & 263.9 \\
\textbf{BCAT-DTF (ours)} & & 75.3 & 90.0 & 92.9& 88.6 & 90.3 & 92.7 & 87.4 & 73.7 & 92.5 & 86.7 & 75.4 & 93.5 & 86.6 & 2871MB & 253.1 \\
\bottomrule
\end{tabular}
}
 \label{tab:inference_models}
\end{table}

As shown in Tables \ref{tab:office31}-\ref{tab:VisDA}, the two inference models proposed in Section \ref{sec:inference} have comparable performance on the four benchmark datasets. This result suggests that the cross-attention feature $Z_{s \rightarrow t}$ with source and target information is close to the self-attention feature $Z_t$ with only target information after training, which may imply that the proposed BCAT method could learn domain-invariant feature representations. In Table \ref{tab:inference_models}, we also compare two inference models with pure BCAT, whose inference model is identical to the training model by utilizing source data, based on ViT and Swin transformer in terms of the classification accuracy on each task and average, occupied GPU memory, and frames per second (FPS) for inference on each image. According to the results, we can see that the proposed KD and DTF inference methods occupy less memory and process images faster than pure BCAT with comparable and even better performance. Those results show that the proposed inference models are both effective and efficient.

\subsection{Visualization of Attention Maps}

\begin{figure} [!htb]
\centering

    \begin{center}
        \rotatebox{90}{\quad Original image}
        \centering
        \subfigure{\includegraphics[width=0.9\textwidth]{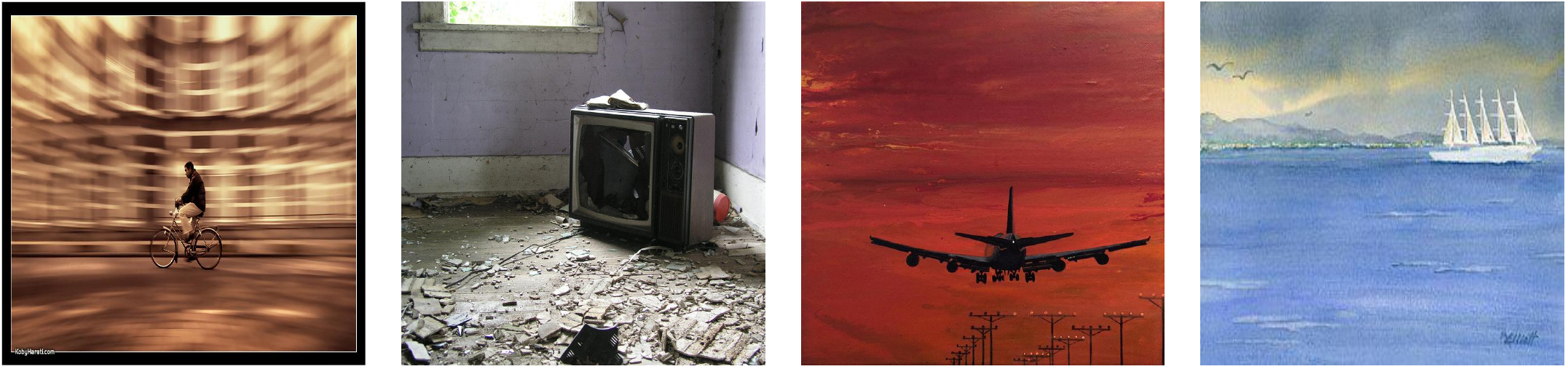}}

        \rotatebox{90}{\quad \; Source-only}
        \centering
        \subfigure{\includegraphics[width=0.9\textwidth]{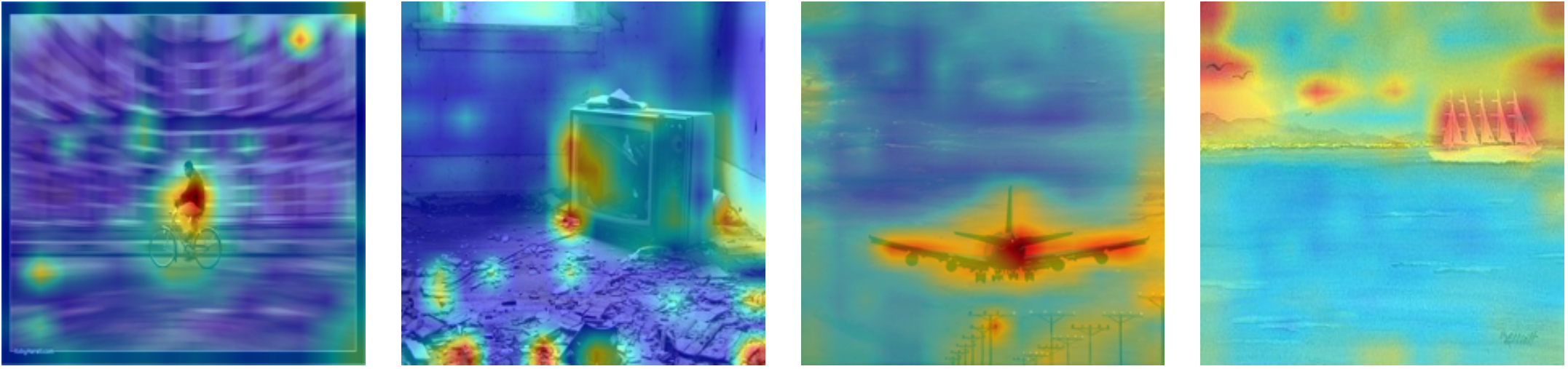}}

        \rotatebox{90}{\qquad \; BCAT}
        \centering
        \subfigure{\includegraphics[width=0.9\textwidth]{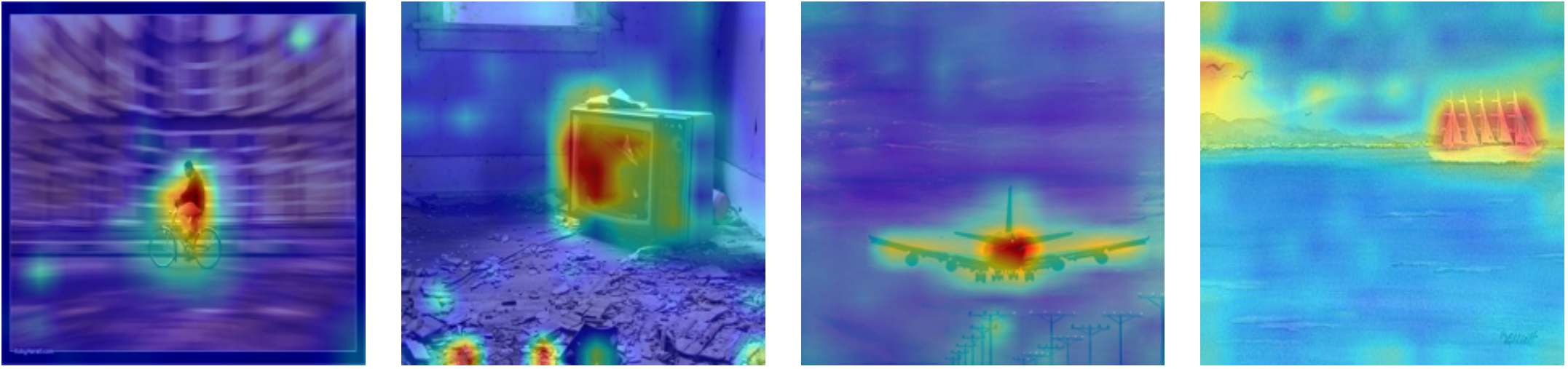}}
    \end{center}
    \caption{Attention maps of images from two classes (i.e., `bicycle' and `TV') in the Office-Home dataset and two classes (i.e., `airplane' and `cruise ship') in the DomainNet dataset. The hotter the color, the higher the attention.}
\label{fig:attention_map}
\end{figure}

According to attention maps shown in Figure \ref{fig:attention_map}, the proposed ViT-based BCAT method can capture important regions more accurately than the `Source-only' baseline and pay less attention to the background. For example, compared with the `Source-only' baseline, the BCAT method focuses almost only on the bicycle in the first image, and it owns more hot areas on the target object than the `Source-only' baseline for the second to fourth images.

\section{Conclusion}

In this paper, we propose the bidirectional cross-attention transformer that is built on the proposed quadruple transformer block. By bidirectionally learning cross-attention between different domains to generate intermediate feature representations, the bidirectional cross-attention and self-attention can be combined to strengthen the domain alignment. Experimental results show that the proposed BCAT outperforms existing state-of-the-art ResNet-based and transformer-based DA methods on four benchmark datasets. In the future study, we are interested in applying BCAT to other computer vision tasks such as semantic segmentation.

%
%
\bibliographystyle{named}
\bibliography{BCAT}
\end{document}